\documentclass[a4paper]{article}
\pdfoutput=1

\usepackage{INTERSPEECH2021}

\usepackage{amsmath,amsfonts,amssymb,bm,mathtools}
\usepackage{booktabs}
\usepackage{hyperref}
\usepackage{url}
\usepackage{appendix}
\usepackage{cite}
\usepackage{multirow}
\usepackage{todonotes}

\title{PnG BERT: Augmented BERT on Phonemes and Graphemes for Neural TTS}
\name{Ye~Jia,~~Heiga~Zen,~~Jonathan~Shen,~~Yu~Zhang,~~Yonghui~Wu}
\address{Google Research}
\email{\{jiaye,heigazen,jonathanasdf,ngyuzh,yonghui\}@google.com}

\begin{document}

\bmdefine{\bO}{O}
\bmdefine{\bC}{C}
\bmdefine{\bc}{c}
\bmdefine{\bo}{o}
\bmdefine{\bW}{W}
\bmdefine{\bmu}{\mu}
\bmdefine{\bQ}{Q}
\bmdefine{\bq}{q}
\bmdefine{\bw}{w}
\bmdefine{\bU}{U}
\bmdefine{\bL}{L}
\bmdefine{\bu}{u}
\bmdefine{\bZero}{0}
\bmdefine{\bI}{I}
\bmdefine{\bR}{R}
\bmdefine{\bP}{P}
\bmdefine{\br}{r}
\bmdefine{\be}{e}
\bmdefine{\bmm}{m}
\bmdefine{\bsigma}{\sigma}
\bmdefine{\bSigma}{\Sigma}
\bmdefine{\bOmega}{\Omega}
\bmdefine{\bomega}{\omega}
\bmdefine{\bS}{S}
\bmdefine{\bA}{A}
\bmdefine{\bC}{C}
\bmdefine{\bM}{M}
\bmdefine{\bg}{g}
\bmdefine{\bs}{s}
\bmdefine{\bpsi}{\psi}
\bmdefine{\bPsi}{\Psi}
\bmdefine{\bphi}{\phi}
\bmdefine{\bPhi}{\Phi}
\bmdefine{\bPi}{\Pi}
\bmdefine{\bpi}{\pi}
\bmdefine{\bLambda}{\Lambda}
\bmdefine{\blambda}{\lambda}
\bmdefine{\bB}{B}
\bmdefine{\bb}{b}
\bmdefine{\bl}{l}
\bmdefine{\bd}{d}
\bmdefine{\bD}{D}
\bmdefine{\bY}{Y}
\bmdefine{\bG}{G}
\bmdefine{\bp}{p}
\bmdefine{\bxi}{\xi}
\bmdefine{\bmeta}{\eta}
\bmdefine{\bzeta}{\zeta}
\bmdefine{\bk}{k}
\bmdefine{\bK}{K}
\bmdefine{\bF}{F}
\bmdefine{\bv}{v}
\bmdefine{\bX}{X}
\bmdefine{\bx}{x}
\bmdefine{\by}{y}
\bmdefine{\bz}{z}
\bmdefine{\bZ}{Z}
\bmdefine{\bcalX}{\mathcal{X}}
\bmdefine{\bH}{H}
\bmdefine{\bh}{h}
\bmdefine{\bcalH}{\mathcal{H}}
\bmdefine{\bV}{V}
\def\diag{\operatorname{diag}}
\def\idiag{\operatorname{diag}^{-1}}
\def\tr{\operatorname{tr}}
\def\vec{\operatorname{vec}}
\def\Gauss{\mathcal{N}}
\def\Qf{\mathcal{Q}}
\def\calM{\mathcal{M}}
\def\Ind{\mathrm{I}}
\def\Err{\mathcal{E}}
\def\Data{\mathcal{D}}
\def\Loss{\mathcal{L}}
\def\ie{\textit{i.e.}, }
\def\eg{\textit{e.g.}, }

\definecolor {GoogleRed}   {rgb}{0.97265625, 0.00390625, 0.00390625}
\definecolor {GoogleBlue}  {rgb}{0.0078125,  0.3984375,  0.78125}
\definecolor {GoogleYellow}{rgb}{0.9453125,  0.70703125, 0.05859375}
\definecolor {GoogleGreen} {rgb}{0.0,        0.57421875, 0.23046875}
\def\GoogleLogo{\textsf{\textcolor{GoogleBlue}{G}\textcolor{GoogleRed}{o}\textcolor{GoogleYellow}{o}\textcolor{GoogleBlue}{g}\textcolor{GoogleGreen}{l}\textcolor{GoogleRed}{e}}}
\def\GoogleAILogo{\GoogleLogo~\textsf{AI}}

\hyphenation{Libri-TTS}
\hyphenation{Libri-Speech}
\hyphenation{Wave-Net}
\hyphenation{Wave-RNN}
\hyphenation{Taco-tron}

\maketitle
\begin{abstract}
This paper introduces \emph{PnG BERT}, a new encoder model for neural TTS. This model is augmented from the original BERT model, by taking both phoneme and grapheme representations of text as input, as well as the word-level alignment between them. It can be pre-trained on a large text corpus in a self-supervised manner, and fine-tuned in a TTS task.
Experimental results show that a neural TTS model using a pre-trained PnG BERT as its encoder yields more natural prosody and more accurate pronunciation than a baseline model using only phoneme input with no pre-training. Subjective side-by-side preference evaluations show that raters have no statistically significant preference between the speech synthesized using a PnG BERT and ground truth recordings from professional speakers.
\end{abstract}

\section{Introduction}
\label{sec:intro}

The advances of neural network-based text-to-speech (TTS) synthesis in the past a few years have closed the gap between synthesized speech and professional human recordings in terms of naturalness \cite{shen2018natural, shen2020non}.
Such neural networks typically consist of an encoder which encodes the input text representation into hidden states, a decoder which decodes spectrogram frames or waveform samples from the hidden states, and an attention or a duration-based upsampler connecting the two together \cite{Char2Wav,oord2016wavenet,arik2017deep,gibiansky2017deep,ping2017deep,wang2017tacotron,shen2018natural,shen2020non,elias2020parallel,li2019neural,ren2019fastspeech,ren2020fastspeech2,weiss2020wave}.

Many early such works take characters of text as input, to demonstrate their capability of being an ``end-to-end'' model (\ie text-analysis-free) \cite{Char2Wav,wang2017tacotron,shen2018natural}. More recent works often use phonemes as input, to achieve better stability and generalize better beyond their training sets \cite{wang2018style,skerry2018towards,jia2018transfer,li2019neural,ren2019fastspeech,ren2020fastspeech2,shen2020non,elias2020parallel,weiss2020wave}.
However, phoneme-based models may suffer from the ambiguity of the representation, such as on homophones. For example, the sentence
``\emph{To cancel the payment, press one; or to continue, two.}''
is a pattern that used frequently by conversational AI agents for call centers.
In the phoneme representation, the trailing ``\emph{..., two.}'' can be easily confused with ``\emph{..., too.}'', which is used more frequently in English.
However, in natural speech, different prosody is expected at the comma positions in these two patterns
-- the former expects a moderate pause at the comma, while having no pause sounds more natural for the latter.

Phoneme and grapheme representations have been combined before in TTS \cite{kastner2019mixing,ming2019feature, chung2019semi, kenter2020improving}, most commonly by concatenating grapheme-based embeddings
to phoneme embeddings \cite{ming2019feature, chung2019semi, kenter2020improving}. However, such approaches face challenges on handling alignment between phonemes and grapheme-based tokens, often require using word-level embeddings (thus a large vocabulary but still with out-of-vocabulary cases) \cite{ming2019feature, chung2019semi}, or discarding a portion of tokens when subword embeddings are used \cite{kenter2020improving}. Another approach is to use a multi-source attention \cite{chung2019semi}, attending to both phoneme sequence and grapheme sequence. However, this approach is restricted to attention-based models only, making it inapplicable to duration-based models. More importantly, it may not fully exploit phoneme-grapheme relationships because of the simple architecture of the attention mechanism.

Self-supervised pre-training on large text corpora, using language model (LM) or masked-language model (MLM) objectives, has proven to be successful in natural language processing in the past decade.
Such pre-training has been applied to TTS for improving the performance both in low-resource scenarios \cite{chung2019semi} and high-resource scenarios \cite{hayashi2019pre, kenter2020improving, xu2020improving}, by using embeddings at subword-level \cite{hayashi2019pre, kenter2020improving}, word-level \cite{chung2019semi}, or sentence-level \cite{hayashi2019pre, xu2020improving}. 
Among the model architectures used for pre-training,
BERT
is one of the most successful ones, and is often adopted for related works in TTS, such as in \cite{hayashi2019pre, kenter2020improving, xu2020improving, zhang2020unified}. 
However, in these works, the pre-training is only performed on graphemes; no phoneme-grapheme relationship is learned during the pre-training.

\urlstyle{same}

This paper introduces \emph{PnG BERT}, an augmented BERT model %
that
can be used as a drop-in replacement for the encoder
in typical neural TTS models, including 
attention-based and duration-based ones. %
PnG BERT can benefit neural TTS models by taking advantages of both phoneme and grapheme representation, as well as by self-supervised pre-training on large text corpora to better understand natural language in its input.
Experimental results show that the use of a pre-trained PnG BERT model can significantly improve the naturalness of synthesized speech, especially in terms of prosody and pronunciations.
Subjective side-by-side preference evaluations show that raters have no statistically significant preference between the  speech synthesized using PnG BERT and ground truth recordings from professional speakers.
Audio samples are available online\footnote{\url{https://google.github.io/tacotron/publications/png_bert/}}.

\section{PnG BERT}
\label{sec:model}

\begin{figure*}[t]
  \centering
  \includegraphics[width=0.96\linewidth]{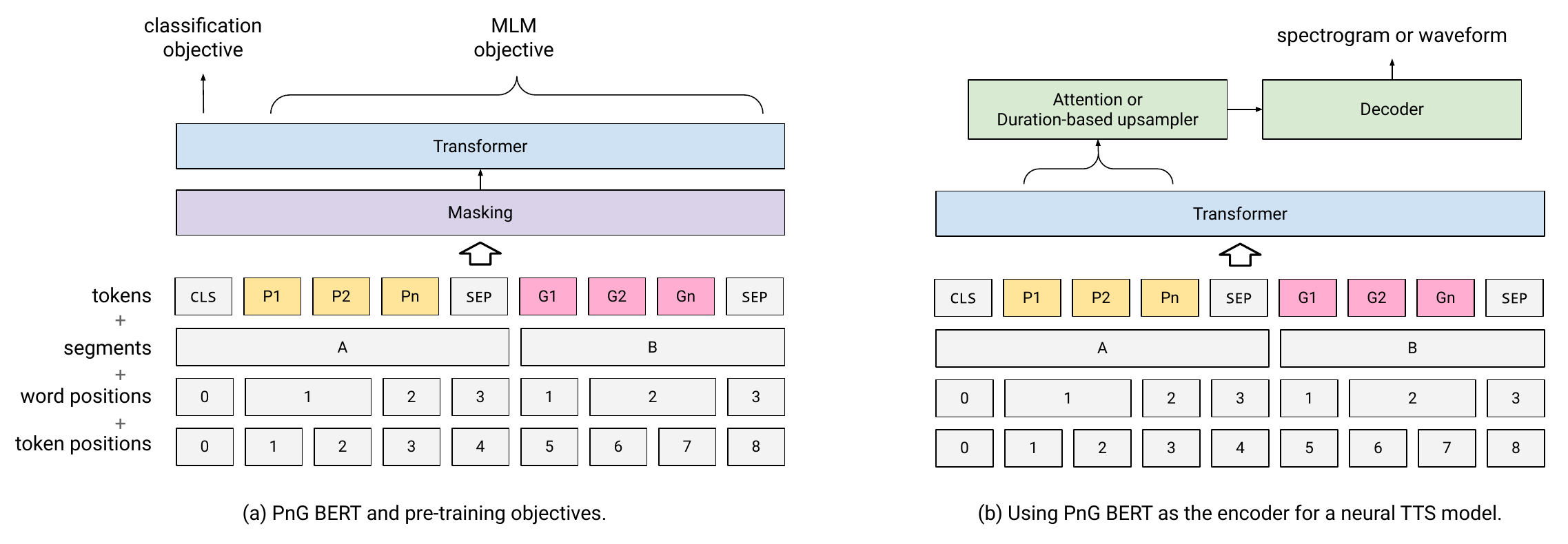}
  \caption{The pre-training and fine-tuning of PnG BERT for neural TTS. Phonemes are displayed in yellow, graphemes in pink.}
  \label{fig:model}
\end{figure*}

The PnG BERT model is illustrated in Figure~\ref{fig:model}. It takes both phonemes and graphemes as input, and can be used directly as an input encoder in a typical neural TTS model. It follows the original BERT architecture \cite{devlin2018bert}, except for differences in the input and the output, and pre-training and fine-tuning procedures. It also bears some similarity to XLM \cite{lample2019cross}, a cross-lingual language model.

\subsection{Background: BERT}

The original BERT model \cite{devlin2018bert} is essentially a Transformer encoder \cite{vaswani2017attention}, pre-trained using a masked language model (MLM) objective on a text corpus. It can take inputs composed of multiple sentences, identified with a segment ID. A special token \texttt{CLS} is prepended to the first segment, which can be used in an extra classification objective in addition to the MLM objective. Another special token \texttt{SEP} is appended to each segment, indicating segment boundaries. The input to BERT is represented as the sum of a token embedding, a segment embedding and a position embedding.

\subsection{Input representation}

The input to PnG BERT is composed of two segments (i.e. sub-sequences), which are two different representations of the same content. The first segment is a phoneme sequence, containing individual IPA phonemes as tokens. The second segment is a grapheme sequence, containing subword units from text as tokens\footnote{The terms ``grapheme'' and ``subword unit'' are considered to be equivalent in the current context, and are used interchangeably in the remainder of this paper.}.
The special tokens \texttt{CLS} and \texttt{SEP} are added in the same way as in the original BERT.
All tokens share the same ID space for embedding lookup and MLM classification.

In addition to the token, segment and position embeddings the original BERT uses, a fourth word-position embedding is used by PnG BERT, providing word-level alignment between phonemes and graphemes. This embedding is implemented with the same sinusoidal functions as the regular position embedding, but using the index of the word each phoneme or grapheme belongs to as the input position. An additional learned linear projection is applied to the sinusoidal signals in order to avoid confusion with the regular position embedding.
All four embeddings are summed together.

Figure~\ref{fig:model} shows a visualized example of such construction of the input.

\subsection{Pre-training}

Similar to the original BERT model, PnG BERT can be pre-trained on a plain text corpus in a self-supervised manner.
Phonemes are obtained using an external grapheme-to-phoneme (G2P) conversion system, while graphemes can be characters, bytes, or obtained using a subword text tokenizer, such as WordPiece \cite{schuster2012japanese}, SentencePiece \cite{kudo2018sentencepiece}, or byte-pair encoding (BPE) \cite{sennrich2015neural}.
Only the MLM objective is used during pre-training.

\subsubsection{Masking strategy}
\label{sec:masking}

The input to PnG BERT consists of two distinct sequences representing essentially the same content. If
random masking was applied as in the original BERT model,
the counterpart of a token masked in one sequence could present in the other sequence. This would make the MLM prediction significantly easier and would reduce the effectiveness of the pre-training.
To avoid this issue, we apply random masking at word-level, consistently masking out phoneme and grapheme tokens belonging to the same word.

We use the same masking categories and ratios used during the original BERT pre-training process: both the phonemes and graphemes for 12\% random words are replaced by \texttt{MSK} tokens; both the phonemes and graphemes for another 1.5\% random words are replaced by random phonemes and graphemes, respectively; both the phonemes and graphemes for another 1.5\% random words are kept unchanged. The MLM loss is computed only on tokens corresponding to these 15\% words.

Alternative 
strategies may be used. In Section~\ref{exp:pretraining} we experimented with increasing the masking ratios of the original random masking. Another choice is to apply masking in phoneme-to-grapheme (P2G) and G2P-like manners, \ie masking out all tokens in one segment and keeping all tokens in the other.

\subsection{Fine-tuning}

The PnG BERT model can be used as an input encoder for typical neural TTS models.
Its weights can be initialized from a pre-trained model, and further fine-tuned during TTS training.
We freeze the weights of the embeddings and the lower Transformer layers, and only fine-tune the higher layers,
to prevent degradation due to the smaller TTS training set and to help the generalization ability of the trained TTS model.
The MLM objective is not used during TTS fine-tuning.

Only the hidden states from the final Transformer layer on the phoneme token positions are passed to the downstream TTS components (e.g., the attention or the duration-based upsampler), as illustrated in Figure~\ref{fig:model}.
Despite these hidden states are only from the phoneme positions, they can carry information from the graphemes because of the self-attention machanism in the PnG BERT model itself.

In the experiments in the next section, we use a NAT TTS model \cite{shen2020non}, and replace the original RNN-based encoder with PnG BERT. We anticipate that 
the PnG BERT model can be applied to other neural TTS model architectures, too.

\section{Experiments}
\label{sec:exp}

To evaluate the performance of PnG BERT, we conducted experiments on a monolingual multi-speaker TTS task. PnG~BERT is used to replace the original RNN-based encoder in NAT \cite{shen2020non}, a duration-based neural TTS model. It produces mel-spectrograms which are converted into the final waveforms using a WaveRNN-based neural vocoder \cite{kalchbrenner2018efficient}.

In all the experiments, the PnG BERT model used 6 Transformer layers with hidden size 512 and 8 attention heads, pre-trained
using the SM3 optimizer \cite{anil2019memory}
with a batch size of 24K for 1M steps and a maximum sequence length of 480 tokens.
The remainder of the TTS models used the same architecture and hyperparameters as in \cite{shen2020non}, except that we trained them with a larger batch size of 512 for 450K steps (same for the standard NAT model used as a baseline). The top 2 Transformer layers in PnG BERT were fine-tuned, with the rest part frozen.

The performance of the TTS models
were evaluated through subjective listening tests on the naturalness, including 5-scale mean opinion score (MOS) tests and 7-scale ($-$3 to 3) side-by-side (SxS) preference tests. 
In the SxS preference tests, 
loudness normalization\footnote{Loudness normalization is critical in SxS preference tests, as raters tend to be biased towards louder samples.} is applied consistently on the all samples in both side.
All subjective tests were conducted using at least 1,000 samples. Each rater was allowed to rate no more than 6 samples in one test.

\subsection{Pre-training performance}
\label{exp:pretraining}

We pre-trained the PnG BERT model on a plain text corpus mined from Wikipedia, containing 131M English sentences. A proprietary text normalization engine \cite{ebden2015kestrel} was used to convert the text into the corresponding phoneme sequences. A SentencePiece model \cite{kudo2018sentencepiece} with a vocabulary of 8,192 tokens was used for tokenizing text into subwords (\ie graphemes).

\begin{figure*}[t]
  \centering
  \includegraphics[width=0.98\linewidth]{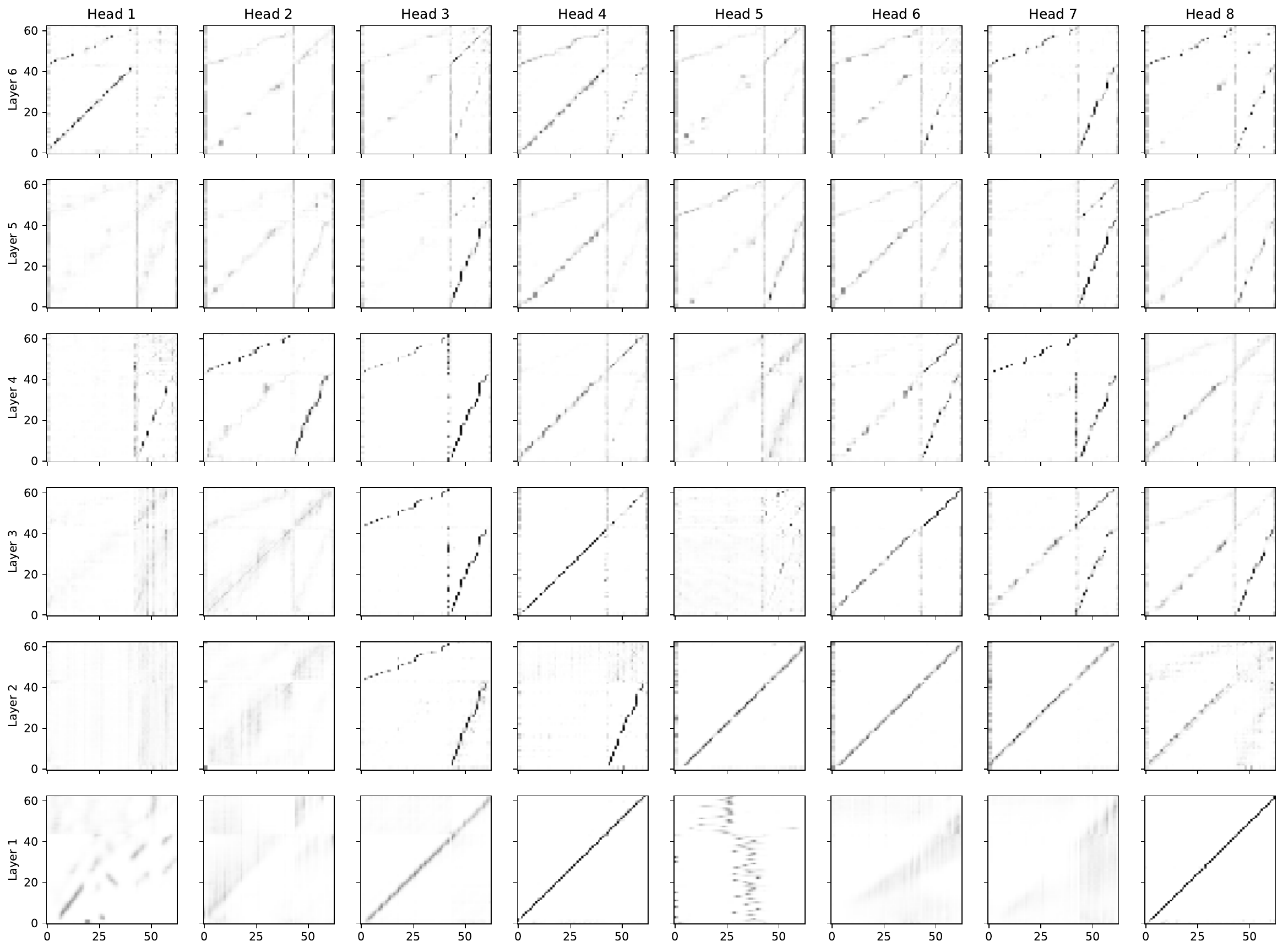}
  \caption{Self-attention probabilities of each layer and head in a PnG BERT without using word-level alignment. The input text is ``To cancel the payment, press one; or to continue, two.'', which is converted into an input sequence composed of a first segment of 42 phonemes and 2 special tokens, and a second segment of 18 graphemes and 1 special token. The trident-shaped distributions in the higher layers indicate that the model learns the alignment between phonemes and graphemes, and encodes information across the two segments. When word-level alignment is provided, such trident-shaped distributions emerge starting from the bottom layer.}
  \label{fig:attention}
\end{figure*}

\begin{table}[tb]
  \caption{Pre-training performance of PnG~BERT, measured by token prediction accuracy using 3 different evaluation-time masking. (MLM: same random masking as training time; G2P / P2G: masking out all phonemes / graphemes, respectively.)}
  \label{tbl:pretraining}
  \centering
  \begin{tabular}{lccc}
    \toprule
    Model                       & MLM &   G2P  & P2G \\
    \midrule
    PnG BERT                    & 67.9 & 50.9 & 33.2 \\
    \quad w/o consistent masking  & 96.8 & 99.1 & 95.2 \\
    \quad\quad w/o word alignment & 94.9 & 41.3 & 39.5 \\
    \bottomrule
  \end{tabular}
\end{table}

We compared multiple pre-training strategies for PnG BERT (Sec.~\ref{sec:masking}). 
For word-level consistent masking, the same masking ratios (15\% in total) as the original BERT were used; otherwise, we doubled the masking ratios (to 30\% in total) since the objective was easier.
As Table~\ref{tbl:pretraining} shows, the word-level consistent masking makes the MLM prediction significantly harder than simply doubling the random masking ratios; it also makes the trained model perform worse in the G2P and P2G evaluations. Providing word-level alignment between phonemes and graphemes (\ie word position embedding) significantly helped the MLM prediction and brought G2P and P2G accuracy to a very high level. However, as shown in the following subsections, these metrics did not 
foretell the performance on the downstream TTS tasks.

Figure~\ref{fig:attention} shows plots of the self-attention probabilities across PnG BERT layers, without providing word-level alignment. The trident-shaped distribution in the higher layers indicates that the model learns the alignment between phonemes and graphemes, and effectively combines information from both.
Such trident-shaped distribution emerges starting from the bottom layer when word-level alignment is provided, suggesting that the PnG BERT learns more effectively in such case, which is consistent with the other experiment results.

\subsection{TTS performance}

\begin{table}[t]
  \centering
  \caption{Performance of the TTS models, measured with 10 speakers (5 male and 5 female) on 3 evaluation sets.}
  \label{tbl:monolingual}
  \begin{tabular}{@{\hspace{0.0em}}l@{\hspace{0.3em}}c@{\hspace{0.4em}}c@{\hspace{0.5em}}c@{\hspace{0.0em}}}
    \toprule
    \multirow{2}{*}{Model}
      & MOS & \multicolumn{2}{c}{SxS (vs Baseline)} \\
      \cmidrule(l{0.3em}r{0.3em}){2-2} \cmidrule(l{0.3em}r{0.3em}){3-4}
      & Generic lines & Hard lines & Questions \\
    \midrule
    NAT (Baseline)                & 4.41 $\pm$ 0.05 & - & - \\
    \midrule
    NAT w/ PnG BERT               & 4.47 $\pm$ 0.05 & 0.28 $\pm$ 0.05 & 0.15 $\pm$ 0.06 \\
    \; w/o consistent masking     & 4.44 $\pm$ 0.05 & 0.14 $\pm$ 0.06 & 0.10 $\pm$ 0.06 \\
    \quad w/o word alignment      & 4.45 $\pm$ 0.05 & 0.15 $\pm$ 0.06 & 0.12 $\pm$ 0.06 \\
    \; w/o pre-training           & 4.30 $\pm$ 0.06 & 0.12 $\pm$ 0.05 & 0.06 $\pm$ 0.05 \\
    \bottomrule
  \end{tabular}
\end{table}

\begin{table}[t]
  \caption{Performance of the TTS models on recordings held-out from training, measured with the same 10 speakers (5 male and 5 female) as in Table~\ref{tbl:monolingual}, each with 100 samples.}
  \label{tbl:monolingual-gt}
  \centering
  \begin{tabular}{lcc}
    \toprule
    Model  & MOS & SxS (vs Ground truth)\\
    \midrule
    NAT (Baseline)    & 4.45 $\pm$ 0.05 & $-0.11 \pm 0.09$ \\
    NAT w/ PnG BERT   & 4.47 $\pm$ 0.05 & $0.02 \pm 0.10$ \\
    \midrule
    Ground truth      & 4.47 $\pm$ 0.05 & - \\
    \bottomrule
  \end{tabular}
\end{table}

We trained multi-speaker TTS models on a proprietary dataset consisting of 243 hours of American English speech, recorded by 31 professional speakers, downsampled to 24 kHz. The amount of recording per speaker varies from 3 to 47 hours.

\subsubsection{Ablation studies}
\label{exp:ablation}

We conducted ablation studies using three evaluation text sets slightly modified from \cite{kenter2020improving}:
a ``generic lines'' set with 1,000 generic lines (the same set as in \cite{shen2020non}), including long inputs (corresponding to up to more than 20 seconds of audio);
a ``hard lines'' set with 286 lines\footnote{Available at \url{https://google.github.io/chive-prosody/chive-bert/dataset}}, containing, e.g., titles and long noun compounds, expected to be hard;
and a ``questions'' set with 300 questions, as questions are prosodically different from statements.
These lines were synthesized using 10 speakers (5 male and 5 female), in a round-robin fashion for the generic lines set (resulting in 1,000 utterances), and a cross-join fashion for the hard lines and questions sets (resulting in 2,860 and 3,000 utterances, respectively).

The results are reported in Table~\ref{tbl:monolingual}. It can be seen that using pre-trained PnG BERT significantly improved the performance of NAT.
The best result was achieved by using word-level consistent masking, providing the word-level phoneme-grapheme alignment, and pre-training on a large text corpus.
When PnG BERT was not pre-trained, the TTS model performed worse than the baseline on the generic line set. Further check revealed that it was due to the trained model not generalizing well to inputs longer than what was seen during training, which is a known limitation of Transformer \cite{dehghani2019universal,hahn2020theoretical}. Nevertheless, it outperformed the baseline on shorter evaluation sets, possibly from the benefit of using both phoneme and grapheme as input.

It is interesting to note that the PnG BERT model with the highest G2P and P2G accuracy during pre-training performed
worse than the ones with significantly lower G2P and P2G accuracy.
This could be because it had the easiest objective during pre-training (reflected by the highest MLM accuracy) and, as a result, did not learn as much as the models with other pre-training strategies did. This finding suggests that the benefit of PnG BERT primarily lies in better natural language understanding, rather than potential improvements on G2P conversion.

\subsubsection{Subjective rater comments}

In the SxS tests, on positive examples, raters comments often mentioned better ``prosody'', ``tone'', ``stress'', ``inflection'', or ``pronunciation''; on negative examples, the comments often mentioned worse ``inflection'', ``stress'', or ``unnatural distortion''. These comments are consistent with the foregoing analysis, confirming that the improvement are primarily from better natural language understanding and thus improved prosody in the synthesized speech.

\subsubsection{Comparison to ground truth recordings}

Lastly, we conducted subjective SxS preference tests against ground truth recordings 
held out from training. We randomly sampled 100 utterances of each of the 10 speakers, and compared the synthesized audios with the ground truth recordings. 
The results in Table~\ref{tbl:monolingual-gt} show that raters had no statistically significant preference between the ground truth recordings from professional speakers and the speech synthesized using PnG BERT.

\section{Conclusions}

We proposed \emph{PnG BERT}, an augmented BERT model that takes both phoneme and grapheme representations of text as its input. We also described a strategy for effectively pre-training it on a large text corpus in a self-supervised manner.
PnG BERT can be directly used as an input encoder for typical neural TTS models. Experimental results showed that PnG BERT can significantly improve the performance of NAT, a state-of-the-art neural TTS model, by producing more natural prosody and more accurate pronunciation. Subjective side-by-side preference evaluation showed that raters had no statistically significant preference between the speech synthesized using PnG BERT and the ground truth recordings from professional speakers.

\section{Acknowledgements}

The authors would like to thank the Google TTS Research team, in particular Tom Kenter
for his support on the evaluation and comprehensive suggestions on the writing.

\bibliographystyle{IEEEtran}
\bibliography{references}

\end{document}